\pdfoutput=1

\documentclass[11pt]{article}

\usepackage[dvipsnames]{xcolor}
\usepackage{naacl2021}

\usepackage{times}
\usepackage{latexsym}
\usepackage{inconsolata}

\usepackage{graphicx}
\usepackage{booktabs}
\usepackage{subfig}
\usepackage[textsize=large]{todonotes}
\usepackage{booktabs}
\usepackage{adjustbox}
\usepackage[ruled]{algorithm2e}
\usepackage{algpseudocode}
\usepackage[inline]{enumitem}
\usepackage{amsmath,amssymb}
\usepackage[font=small,skip=5pt]{caption}
\usepackage{tabularx} %
\usepackage{longtable}
\usepackage{url}
\usepackage{cleveref}
\usepackage{enumitem}
\usepackage[compact]{titlesec}
\usepackage{multirow}
\usepackage{wasysym}
\usepackage{wrapfig}
\usepackage{xspace}
\usepackage{pgfplots, pgfplotstable}
\pgfplotsset{compat=1.16}
\usepackage{filecontents}
\usepackage{subfig}
\usepackage{xcolor}
\usepackage[title]{appendix}

\usepackage{microtype}

\newif\ifcomment
\commenttrue

\newcommand{\R}{\ensuremath{\mathbb{R}}}
\newcommand{\Upool}{\ensuremath{\mathcal{U}}}
\newcommand{\Lpool}{\ensuremath{\mathcal{L}}}
\newcommand{\Model}{\ensuremath{\mathcal{M}}}
\newcommand{\Criterion}{\ensuremath{\mathcal{C}}}

\newcommand{\dpp}{DPP\xspace}

\newcommand{\vect}[1]{\ensuremath{\mathbf{#1}}}

\newcommand{\refsec}[1]{\textsection\ref{#1}}
\newcommand{\reftab}[1]{Table~\ref{#1}}
\newcommand{\reffig}[1]{Figure~\ref{#1}}
\newcommand{\refalgo}[1]{Algorithm~\ref{#1}}

\newcommand{\numstd}[2]{\ensuremath{#1_{\pm #2}}}
\newcommand{\numstdbf}[2]{\ensuremath{\mathbf{#1}_{\pm #2}}}

\pgfplotstableset{col sep=comma}
\usepgfplotslibrary{fillbetween}

\title{Diversity-Aware Batch Active Learning for Dependency Parsing}

\author{
  Tianze Shi\thanks{~~Work done during an internship at Bloomberg L.P.}\\
  Cornell University \\
  {\tt tianze@cs.cornell.edu} \\\And
  Adrian Benton\\
  Bloomberg L.P.\\
  {\tt abenton10@bloomberg.net}\\\AND
  Igor Malioutov\\
  Bloomberg L.P. \\
  {\tt imalioutov@bloomberg.net} \\\And
  Ozan {\.I}rsoy\\
  Bloomberg L.P. \\
  {\tt oirsoy@bloomberg.net} \\
}

\date{}

\begin{document}

\maketitle

\begin{abstract}

While the predictive performance of modern statistical dependency parsers relies heavily
on the availability of expensive expert-annotated treebank data,
not all annotations contribute equally to the training of the parsers.
In this paper, we attempt to reduce the
number of labeled examples needed to train a strong dependency parser using
batch active learning (AL).
In particular, we investigate whether enforcing
diversity in the sampled batches,
using determinantal point processes (DPPs),
can
improve over their diversity-agnostic counterparts.
Simulation experiments on an English newswire corpus show that
selecting diverse batches with DPPs is superior to strong
selection strategies that do not enforce batch diversity,
especially during the initial stages of the learning process.
Additionally, our diversity-aware strategy is robust
under a corpus duplication setting,
where diversity-agnostic sampling strategies
exhibit significant degradation.

\end{abstract}

\section{Introduction}
\label{sec:introduction}

Though critical to parser training,
data annotations for dependency parsing are
both expensive and time-consuming to obtain.
Syntactic analysis requires linguistic expertise
and even after extensive training,
data annotation can still be burdensome.
The Penn Treebank project \citep{marcus+93} reports that
after two months of training,
the annotators average $750$ tokens per hour on the bracketing task;
the Prague Dependency Treebank \citep{bohmova+03}
cost over \$$600{,}000$ and required $5$ years to annotate
roughly $90{,}000$ sentences (over \$$5$ per sentence).
These high annotation costs present a significant challenge
to developing accurate dependency parsers
for under-resourced languages and domains.

Active learning \citep[AL;][]{settles2009active} is
a promising
technique
to reduce
the annotation effort required
to train a strong dependency parser by intelligently selecting samples to annotate
such that the return of each annotator hour is as high as possible.
Popular selection strategies, such as uncertainty sampling,
associate each instance with a \emph{quality} measure
based on the uncertainty or confidence level of the current parser,
and higher-quality instances are selected for annotation.
We focus on \emph{batch mode} AL,
since it is generally more efficient
for annotators to label in bulk.
While early work in AL for parsing \citep{tang+02,hwa00,hwa04}
cautions against using individually-computed quality measures in the batch setting,
more recent work demonstrates empirical success \citep[e.g.,][]{li2016active}
without explicitly handling intra-batch \emph{diversity}.
In this paper, we explore whether a diversity-aware approach can improve
the state of the art in AL for dependency parsing.
Specifically, we consider samples drawn from
determinantal point processes (\dpp{}s) as a query strategy to
select batches of high-quality, yet dissimilar instances \cite{kulesza2012determinantal}.

In this paper, we
(1) propose a diversity-aware batch AL query strategy for dependency parsing
compatible with existing selection strategies,
(2) empirically study three AL strategies with and without diversity factors,
and (3) find that diversity-aware selection strategies
are superior to their diversity-agnostic counterparts, especially during the early stages of the learning process, in simulation experiments on an English newswire corpus.  This is critical in low-budget AL settings, which
we further confirm in a corpus duplication setting.\footnote{
Our code is publicly available at \url{https://github.com/tzshi/dpp-al-parsing-naacl21}.
}

\section{Active Learning for Dependency Parsing}
\label{sec:al_for_deppar}

\subsection{Dependency Parsing}
\label{sec:dependency_parsing}

Dependency parsing \cite{kubler+08} aims
to find the syntactic dependency structure, $y$,
given a length-$n$ input sentence $x=x_1, x_2, \ldots, x_n$,
where $y$ is a set of $n$ arcs over the tokens
and the dummy root symbol $x_0$,
and each arc $(h, m)\in y$
specifies the head, $h$, and modifier word, $m$.\footnote{
For clarity, here we describe unlabeled parsing.
In our experiments, we train labeled dependency parsers, which additionally predict a dependency relation label $l$ for each arc.
}
In this work,
we adopt the conceptually-simple
edge-factored deep biaffine dependency parser \cite{dozat-manning17},
which
is competitive with the state of the art in terms of accuracy,
The parser assigns a locally-normalized attachment probability
$P_\text{att}(\text{head}(m)=h~|~x)$
to each attachment candidate pair $(h, m)$ based on a biaffine scoring function.
Refer to Appendix \ref{app:parser} for architecture details.
We define the score of the candidate parse tree $s(y\mid x)$
as $\sum_{(h,m)\in y} \log P_\text{att}(\text{head}(m)=h\mid x)$.
The decoder finds the best scoring $\hat{y}$
among all valid trees $\mathcal{Y}(x)$:
$\hat{y}=\arg\max_{y\in \mathcal{Y}(x)}s(y\mid x)$.

\subsection{Active Learning (AL)}
\label{sec:active_learning}

We consider the pool-based batch AL scenario
where we assume a large collection of unlabeled instances $\Upool$
from which we sample a small subset at a time
to annotate after each round
to form an expanding labeled training set $\Lpool$ \cite{lewis1994sequential}.
We use the superscript $i$ to denote
the pool of instances $\Upool^i$ and $\Lpool^i$
after the $i$-th round.
$\Lpool^0$ is a small set of seed labeled instances to initiate the process.
Each iteration starts with training a model $\Model^i$
based on $\Lpool^i$.
Next, all unlabeled data instances in $\Upool^i$
are parsed by $\Model^i$
and we select a batch $\Upool'$ to annotate
based on some criterion $\Upool'=\Criterion(\Model^i, \Upool^i)$.
The resulting labeled subset $\Lpool'$
is added to $\Lpool^{i+1}=\Lpool^i\bigcup \Lpool'$
and $\Upool^{i+1}=\Upool^i - \Upool'$.

The definition of the selection criterion $\Criterion$
is critical.
A typical strategy associates each unlabeled instance $\Upool_i$
with a quality measure $q_i$
based on, for example,
the model uncertainty level
when parsing $\Upool_i$.
A \emph{diversity-agnostic} criterion sorts
all unlabeled instances by their quality measures
and takes the top-$k$ as $\Upool'$ for a budget $k$.

\subsection{Quality Measures}
\label{subsec:quality_measures}

We consider three commonly-used quality measures
adapted to the task of dependency parsing,
including uncertainty sampling,
Bayesian active learning,
and a representativeness-based strategy.

\paragraph{Average Marginal Probability (AMP)}
measures parser uncertainty \citep{li2016active}:
\begin{center}
$
\texttt{AMP} = 1 - \frac{1}{n} \sum_{(\hat{h},m) \in \hat{y}} {P_\text{mar}(\text{head}(m)=\hat{h} \mid x)},
$
\end{center}
where $P_\text{mar}$ is the marginal attachment probability
\begin{center}
$
P_\text{mar}(\text{head}(m)=h\mid x) = \sum_{(h,m) \in y} P(y\mid x),
$
\end{center}
and $P(y\mid x) = \frac{\exp(s(y \mid x))}{\sum_{y'\in \mathcal{Y}(x)}{\exp(s(y'\mid x))}}$.
The marginal probabilities can be derived efficiently using
Kirchhoff's theorem \citep{tutte84,koo+07}.

\paragraph{Bayesian Active Learning by Disagreement (BALD)}
measures the mutual information between the model parameters
and the predictions.
We adopt the Monte Carlo dropout-based variant \citep{gal+17,siddhant-lipton18}
and measure the disagreement among predictions from a neural model with $K$ different dropout masks, which has been applied to
active learning in NLP.
We adapt BALD to dependency parsing
by aggregating disagreement at a token level:
\begin{center}
$
\texttt{BALD} = 1 - \frac{1}{n} \sum_{m} \frac{\text{count}(\text{mode}(h_m^1, \ldots, h_m^K))}{K},
$
\end{center}
where $h_m^k$ denotes that $(h_m^k, m)$ appears
in the prediction given by the $k$-th model.

\paragraph{Information Density (ID)}
mitigates the tendency of uncertainty sampling to favor outliers
by weighing examples by how \emph{representative} they are of the entire dataset \citep{settles2008analysis}:
\begin{center}
$
\texttt{ID} = \texttt{AMP} \times \left( \frac{1}{|\Upool|} \sum_{x'\in \Upool} \text{sim}_\text{cos}(x, x') \right),
$
\end{center}
where cosine similarity is computed
from the averaged contextualized features (\refsec{sec:diversity_features}).

\subsection{Learning from Partial Annotations}
\label{sec:partial_annotations}

We follow \citet{li2016active}
and select tokens to annotate their heads instead of annotating full sentences.
We first pick the most informative sentences
and then choose $p\%$ tokens from them
based on the token-level versions of the quality measures
(e.g., marginal probability instead of AMP).

\section{Selecting Diverse Samples}
\label{sec:diversity}

Near-duplicate examples are common
in real-world data \citep{broder+97,manku+07},
but they provide overlapping utility to model training.
In the extreme case, with a diversity-agnostic strategy
for active learning,
identical examples will be selected/excluded
at the same time \citep{hwa04}.
To address this issue and to best utilize the annotation budget,
it is important to consider diversity.
We adapt \citet{biyik2019batch}
to explicitly model diversity using determinantal point processes  (\dpp{}s).

\subsection{Determinantal Point Processes}
\label{sec:dpp}

A \dpp{} defines a probability distribution
over subsets of some ground set of elements \cite{kulesza2012thesis}.
In AL,
the ground set is the unlabeled pool
$\Upool$
and a subset corresponds to a batch of
instances $\Upool'$ drawn from $\Upool$.
\dpp{}s provide an explicit mechanism to
ensure high-quality yet diverse sample selection
by modeling both the quality measures
and the similarities among examples.
We adopt the $L$-ensemble representation of \dpp{}s
using the quality-diversity decomposition
\citep{kulesza2012determinantal}
and parameterize the matrix $L$ as
$L_{ij}=q_i \phi_i \phi_j^T q_j$,
where each $q_i\in \R$ is the quality measure for $\Upool_i$
and each $\phi_i\in \R^{1\times d}$ is
a $d$-dimensional vector representation of $\Upool_i$,
which we refer to as $\Upool_i$'s \emph{diversity features}.\footnote{
Although certain applications of \dpp{}s
may learn $q$ and $\phi$ representations from supervision,
we define $q$ and $\phi$ \emph{a priori}, since acquiring supervision in AL is, by definition, expensive.
}
The probability of selecting a
batch $B$
is given by
$P(B \subseteq \Upool) \propto \det(L_{B})$,
where $\det(\cdot)$ calculates the determinant
and $L_B$ is the submatrix of $L$
indexed by elements in $B$.
\dpp{}s place high probability on
diverse subsets of high-quality items.
Intuitively, the determinant of $L_B$ corresponds to
the volume spanned by the set of vectors $\{q_i\phi_i\mid i\in B\}$,
and subsets with larger $q$ values and orthogonal $\phi$ vectors
span larger volumes than
those with smaller $q$ values or similar $\phi$ vectors.
We follow \citet{kulesza2012thesis} and adapt their greedy algorithm for finding the approximate mode
$\arg\max_B P(B\subseteq \Upool)$.
This algorithm is reproduced in \refalgo{algo:dpp-map} in the appendix.

\subsection{Diversity Features}
\label{sec:diversity_features}

We consider two possibilities for
the diversity features $\phi$.
Each feature vector is unit-normalized.
\paragraph{Averaged Contextualized Features}
are defined as
$\frac{1}{n}\sum_i \vect{x}_i$, where $\vect{x}_i$ is a contextualized vector of $x_i$
from the feature extractor used by the parser.
By this definition, we consider the instances to be similar to each other
when the neural feature extractor returns similar features
such that the parser is likely to predict similar structures for
these instances.

\paragraph{Predicted Subgraph Counts}
explicitly represent the predicted tree structure.
To balance richness and sparsity,
we count the labeled but unlexicalized subgraph formed by
the grandparent, the parent and the token itself.
Specifically, for each token $m$,
we can extract a subgraph denoted by $(r_1, r_2)$,
assuming the predicted dependency relation
between its grandparent $g$ and its parent $h$ is $r_1$,
and the relation between $h$ and $m$ is $r_2$.
The parse tree for a length-$n$ sentence contains $n$ such subgraphs.
We apply tf-idf weighting
to discount the influence from frequent subgraphs.

\section{Experiments and Results}
\label{sec:experiments}

\paragraph{Dataset}

We use the Revised English News Text Treebank\footnote{
    \url{https://catalog.ldc.upenn.edu/LDC2015T13}
} \citep{bies+15}
converted to Universal Dependencies 2.0 using the conversion tool included in Stanford Parser \citep{manning+14} version 4.0.0.
We use sections 02-21 for training, 22 for development
and 23 for test.

\paragraph{Setting}
We perform experiments by simulating the annotation process
using treebank data.
We sample $128$ sentences uniformly
for the initial labeled pool
and each following round selects
$500$ tokens for partial annotation.
We run each setting five times using different
random initializations and report the means and standard deviations
of the labeled attachment scores (LAS).
Appendix \ref{app:uas} has
unlabeled attachment score (UAS) results.

\paragraph{Baselines}

While we construct our own baselines for self-contained comparisons,
the diversity-agnostic AMP (w/o DPP) largely replicates
the state-of-the-art selection strategy of \citet{li2016active}.

\paragraph{Implementation}

We finetune a pretrained multilingual XLM-RoBERTa base model \citep{conneau+20}
as our feature extractor.\footnote{
To construct the averaged contextualized features,
we also use the fine-tuned feature extractor.
In our preliminary experiments, we have tried freezing the feature extractors, but this variant did not perform as well.
}
See Appendix \ref{app:hyperparameters} for implementation details.

\paragraph{Main Results}

\begin{table}[]
    \small
    \begin{tabular}{@{\hspace{3pt}}l@{\hspace{2pt}}|c@{\hspace{3pt}}c|c@{\hspace{3pt}}c@{\hspace{3pt}}}
    \toprule
     Batch & \multicolumn{2}{c|}{$5$} & \multicolumn{2}{c}{$10$} \\
    \midrule
     Strategy & w/o DPP   & w/ DPP   & w/o DPP   & w/ DPP   \\
    \midrule
     Random & \numstd{85.68}{.26}  & \numstdbf{86.61}{.28}
            & \numstd{87.84}{.26}  & \numstdbf{88.55}{.23} \\
     AMP    & \numstd{85.98}{.22}  & \numstdbf{86.77}{.43}
            & \numstd{88.80}{.18}  & \numstdbf{89.23}{.29} \\
     BALD   & \numstd{86.24}{.40}  & \numstdbf{86.86}{.31}
            & \numstd{88.66}{.36}  & \numstdbf{89.03}{.10} \\
     ID     & \numstdbf{86.68}{.26}  & \numstd{86.56}{.24}
            & \numstd{88.96}{.20}  & \numstdbf{89.06}{.16} \\
    \bottomrule
    \end{tabular}
    \caption{LAS after $5$ and $10$ rounds of annotation for strategies with and without modeling diversity through DPP.}
    \label{tab:main-results}
\end{table}

\reftab{tab:main-results} compares LAS
after $5$ and $10$ rounds of annotation.
Our dependency parser reaches $95.64$ UAS and $94.06$ LAS,
when trained with the full dataset (more than one million tokens).
Training data collected from $30$ annotation rounds ($\approx 17{,}500$ tokens)
correspond to roughly $2\%$ of the full dataset,
but already support an LAS of up to $92$ through AL.
We find that diversity-aware strategies generally improve over
their diversity-agnostic counterparts.
Even for a random selection strategy, ensuring diversity with a DPP
is superior to simple random selection.
With AMP and BALD,
our diversity-aware strategy sees a larger improvement
earlier in the learning process.
ID models representativeness of instances, and our diversity-aware strategy
adds less utility compared with
other quality measures,
although we do notice a large improvement after the first annotation round for ID:
\numstd{82.40}{.48} vs. \numstd{83.36}{.54} (w/ DPP) -- a similar trend to AMP and BALD, but at an earlier stage of AL.

\paragraph{Experiments with Different Diversity Features}

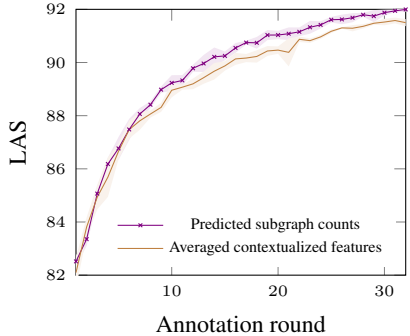
\begin{figure}
    \centering
    \begin{tikzpicture}[]
        \begin{axis}[
            width=0.37*\textwidth,
            grid=none,
            ylabel=LAS,
            xlabel=Annotation round,
            xmin=1,
            xmax=32,
            ymin=82,
            ymax=92,
            legend pos=south east,
            legend style={draw=none, font=\tiny, row sep=-1pt},
            yticklabel style = {font=\tiny},
            xticklabel style = {font=\tiny},
            ylabel style = {font=\small},
            xlabel style = {font=\small},
        ]

        \addplot [violet,mark=x,mark size=1pt] table [x=idx, y=ampdpp] {data.dat};
        \addplot [brown,mark=.,mark size=1pt] table [x=idx, y=ampdppavg] {data.dat};

        \addplot [name path=upper,draw=none] table[x=idx,y expr=\thisrow{ampdpp}+\thisrow{ampdpperr}] {data.dat};
        \addplot [name path=lower,draw=none] table[x=idx,y expr=\thisrow{ampdpp}-\thisrow{ampdpperr}] {data.dat};
        \addplot [fill=violet!10] fill between[of=upper and lower];

        \addplot [name path=upper,draw=none] table[x=idx,y expr=\thisrow{ampdppavg}+\thisrow{ampdppavgerr}] {data.dat};
        \addplot [name path=lower,draw=none] table[x=idx,y expr=\thisrow{ampdppavg}-\thisrow{ampdppavgerr}] {data.dat};
        \addplot [fill=brown!10] fill between[of=upper and lower];

        \legend{Predicted subgraph counts, Averaged contextualized features}

        \end{axis}
    \end{tikzpicture}
\caption{
    Learning curves for our DPP-based diversity-aware selection strategies,
    comparing predicted subgraph counts versus
    averaged contextualized features as diversity features.
    Both use AMP as their quality measures.
}
\label{fig:features}
\end{figure}

\reffig{fig:features} compares our two definitions of diversity features,
and we find that predicted subgraph counts provide stronger performance
than that of averaged contextualized features.
We hypothesize this is due to the fact that
the subgraph counts represent structures more explicitly,
thus they are more useful in maintaining structural diversity in AL.

\paragraph{Intra-Batch Diversity}

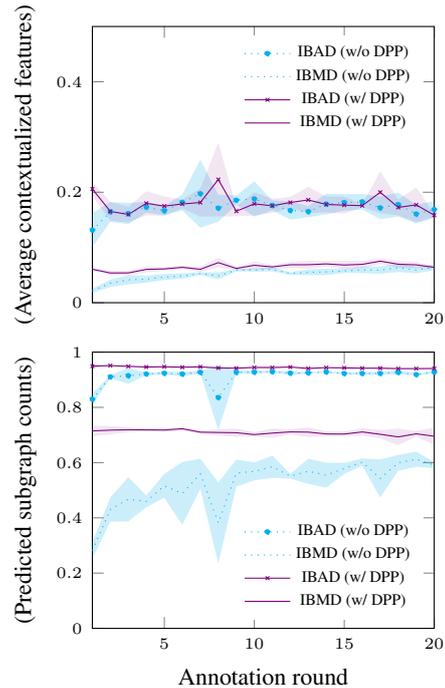
\begin{figure}
    \centering
    \begin{tikzpicture}[]
        \begin{axis}[
            width=0.38*\textwidth,
            grid=none,
            xmin=1,
            xmax=20,
            ymin=0,
            ymax=0.5,
            ylabel=(Average contextualized features),
            legend pos=north east,
            legend style={draw=none, font=\tiny, row sep=-1pt},
            yticklabel style = {font=\tiny},
            xticklabel style = {font=\tiny},
            ylabel style = {font=\small},
            xlabel style = {font=\small},
        ]

        \addplot [cyan,mark=*,dotted,mark size=1pt] table [x=idx, y=amp-avg-ad] {data-dis.dat};
        \addplot [cyan,mark=.,dotted,mark size=1pt] table [x=idx, y=amp-avg-md] {data-dis.dat};
        \addplot [violet,mark=x,mark size=1pt] table [x=idx, y=amp-dpp-avg-ad] {data-dis.dat};
        \addplot [violet,mark=.,mark size=1pt] table [x=idx, y=amp-dpp-avg-md] {data-dis.dat};

        \addplot [name path=upper,draw=none] table[x=idx,y expr=\thisrow{amp-avg-ad}+\thisrow{amp-avg-ad-err}] {data-dis.dat};
        \addplot [name path=lower,draw=none] table[x=idx,y expr=\thisrow{amp-avg-ad}-\thisrow{amp-avg-ad-err}] {data-dis.dat};
        \addplot [fill=cyan,fill opacity=0.2] fill between[of=upper and lower];

        \addplot [name path=upper,draw=none] table[x=idx,y expr=\thisrow{amp-avg-md}+\thisrow{amp-avg-md-err}] {data-dis.dat};
        \addplot [name path=lower,draw=none] table[x=idx,y expr=\thisrow{amp-avg-md}-\thisrow{amp-avg-md-err}] {data-dis.dat};
        \addplot [fill=cyan,fill opacity=0.2] fill between[of=upper and lower];

        \addplot [name path=upper,draw=none] table[x=idx,y expr=\thisrow{amp-dpp-avg-ad}+\thisrow{amp-dpp-avg-ad-err}] {data-dis.dat};
        \addplot [name path=lower,draw=none] table[x=idx,y expr=\thisrow{amp-dpp-avg-ad}-\thisrow{amp-dpp-avg-ad-err}] {data-dis.dat};
        \addplot [fill=violet,fill opacity=0.1] fill between[of=upper and lower];

        \addplot [name path=upper,draw=none] table[x=idx,y expr=\thisrow{amp-dpp-avg-md}+\thisrow{amp-dpp-avg-md-err}] {data-dis.dat};
        \addplot [name path=lower,draw=none] table[x=idx,y expr=\thisrow{amp-dpp-avg-md}-\thisrow{amp-dpp-avg-md-err}] {data-dis.dat};
        \addplot [fill=violet,fill opacity=0.1] fill between[of=upper and lower];

        \legend{IBAD (w/o DPP), IBMD (w/o DPP), IBAD (w/ DPP), IBMD (w/ DPP)}

        \end{axis}
    \end{tikzpicture}

    \begin{tikzpicture}[]
        \begin{axis}[
            width=0.38*\textwidth,
            grid=none,
            xmin=1,
            xmax=20,
            ymin=0,
            ymax=1.0,
            ylabel=(Predicted subgraph counts),
            xlabel=Annotation round,
            legend pos=south east,
            legend style={draw=none, font=\tiny, row sep=-1pt},
            yticklabel style = {font=\tiny},
            xticklabel style = {font=\tiny},
            ylabel style = {font=\small},
            xlabel style = {font=\small},
        ]

        \addplot [cyan,mark=*,dotted,mark size=1pt] table [x=idx, y=amp-crafted-ad] {data-dis.dat};
        \addplot [cyan,mark=.,dotted,mark size=1pt] table [x=idx, y=amp-crafted-md] {data-dis.dat};
        \addplot [violet,mark=x,mark size=1pt] table [x=idx, y=amp-dpp-crafted-ad] {data-dis.dat};
        \addplot [violet,mark=.,mark size=1pt] table [x=idx, y=amp-dpp-crafted-md] {data-dis.dat};

        \addplot [name path=upper,draw=none] table[x=idx,y expr=\thisrow{amp-crafted-ad}+\thisrow{amp-crafted-ad-err}] {data-dis.dat};
        \addplot [name path=lower,draw=none] table[x=idx,y expr=\thisrow{amp-crafted-ad}-\thisrow{amp-crafted-ad-err}] {data-dis.dat};
        \addplot [fill=cyan,fill opacity=0.2] fill between[of=upper and lower];

        \addplot [name path=upper,draw=none] table[x=idx,y expr=\thisrow{amp-crafted-md}+\thisrow{amp-crafted-md-err}] {data-dis.dat};
        \addplot [name path=lower,draw=none] table[x=idx,y expr=\thisrow{amp-crafted-md}-\thisrow{amp-crafted-md-err}] {data-dis.dat};
        \addplot [fill=cyan,fill opacity=0.2] fill between[of=upper and lower];

        \addplot [name path=upper,draw=none] table[x=idx,y expr=\thisrow{amp-dpp-crafted-ad}+\thisrow{amp-dpp-crafted-ad-err}] {data-dis.dat};
        \addplot [name path=lower,draw=none] table[x=idx,y expr=\thisrow{amp-dpp-crafted-ad}-\thisrow{amp-dpp-crafted-ad-err}] {data-dis.dat};
        \addplot [fill=violet,fill opacity=0.1] fill between[of=upper and lower];

        \addplot [name path=upper,draw=none] table[x=idx,y expr=\thisrow{amp-dpp-crafted-md}+\thisrow{amp-dpp-crafted-md-err}] {data-dis.dat};
        \addplot [name path=lower,draw=none] table[x=idx,y expr=\thisrow{amp-dpp-crafted-md}-\thisrow{amp-dpp-crafted-md-err}] {data-dis.dat};
        \addplot [fill=violet,fill opacity=0.1] fill between[of=upper and lower];

        \legend{IBAD (w/o DPP), IBMD (w/o DPP), IBAD (w/ DPP), IBMD (w/ DPP)}

        \end{axis}
    \end{tikzpicture}
\caption{
Intra-batch average distance (IBAD) and intra-batch minimal distance (IBMD) measures
comparing diversity-agnostic and diversity-aware AMP-based sample selection strategies.
The distances are derived from averaged contextualized features (top)
and predicted subgraph counts (bottom).
A higher value indicates better intra-batch diversity.
}
\label{fig:dis}
\end{figure}

To quantify intra-batch diversity among the set of sentences $B$ picked
by the selection strategies,
we adapt the measures used by \citet{chen+18a} and define
intra-batch average distance (IBAD) and
intra-batch minimal distance (IBMD) as follows:
\begin{align*}
    \text{IBAD} &= \underset{i,j\in B,i\neq j}{\mathrm{mean}}(1 - \text{sim}_\text{cos}(i,j)),\\
    \text{IBMD} &= \underset{i\in B}{\mathrm{mean}}\underset{j\in B,i\neq j}{\mathrm{min}}(1 - \text{sim}_\text{cos}(i,j)).
\end{align*}
A higher value on these measures indicates better intra-batch diversity.
\reffig{fig:dis} compares diversity-agnostic and diversity-aware
sampling strategies using the two different diversity features.
We confirm that DPPs indeed promote diverse samples in the selected batches,
while intra-batch diversity naturally increases
even for the diversity-agnostic strategies.
Additionally, we observe that the benefits of DPPs
are more prominent when using predicted subgraph counts
compared with averaged contextualized features.
This can help explain the relative success of the former diversity features.

\paragraph{Corpus Duplication Setting}

\begin{figure*}
    \centering
\subfloat{
    \begin{tikzpicture}[]
        \begin{axis}[
            width=0.37*\textwidth,
            grid=none,
            ylabel=LAS,
            xmin=1,
            xmax=32,
            ymin=82,
            ymax=92,
            legend pos=south east,
            legend style={draw=none, font=\tiny, row sep=-1pt},
            yticklabel style = {font=\tiny},
            xticklabel style = {font=\tiny},
            ylabel style = {font=\small},
        ]

        \addplot [gray,mark=*,mark size=1pt] table [x=idx, y=rand] {data.dat};
        \addplot [cyan,mark=.,mark size=1pt] table [x=idx, y=amp] {data.dat};
        \addplot [violet,mark=x,mark size=1pt] table [x=idx, y=ampdpp] {data.dat};
        \addplot [cyan,mark=.,mark size=1pt, dotted, thick, mark options=solid] table [x=idx, y=ampdup] {data.dat};
        \addplot [violet,mark=x,mark size=1pt, dotted, mark options=solid] table [x=idx, y=ampdppdup] {data.dat};

        \addplot [name path=upper,draw=none] table[x=idx,y expr=\thisrow{rand}+\thisrow{randerr}] {data.dat};
        \addplot [name path=lower,draw=none] table[x=idx,y expr=\thisrow{rand}-\thisrow{randerr}] {data.dat};
        \addplot [fill=gray!20] fill between[of=upper and lower];

        \addplot [name path=upper,draw=none] table[x=idx,y expr=\thisrow{amp}+\thisrow{amperr}] {data.dat};
        \addplot [name path=lower,draw=none] table[x=idx,y expr=\thisrow{amp}-\thisrow{amperr}] {data.dat};
        \addplot [fill=cyan!10] fill between[of=upper and lower];

        \addplot [name path=upper,draw=none] table[x=idx,y expr=\thisrow{ampdpp}+\thisrow{ampdpperr}] {data.dat};
        \addplot [name path=lower,draw=none] table[x=idx,y expr=\thisrow{ampdpp}-\thisrow{ampdpperr}] {data.dat};
        \addplot [fill=violet!10] fill between[of=upper and lower];

        \addplot [name path=upper,draw=none] table[x=idx,y expr=\thisrow{ampdup}+\thisrow{ampduperr}] {data.dat};
        \addplot [name path=lower,draw=none] table[x=idx,y expr=\thisrow{ampdup}-\thisrow{ampduperr}] {data.dat};
        \addplot [fill=cyan!10] fill between[of=upper and lower];

        \legend{Random,AMP,AMP w/ DPP,AMP (dup),AMP w/ DPP (dup)}

        \end{axis}
    \end{tikzpicture}
}
\subfloat{
    \begin{tikzpicture}[]
        \begin{axis}[
            width=0.37*\textwidth,
            grid=none,
            xmin=1,
            xmax=32,
            ymin=82,
            ymax=92,
            legend pos=south east,
            legend style={draw=none, font=\tiny, row sep=-1pt},
            yticklabel style = {font=\tiny},
            xticklabel style = {font=\tiny},
        ]

        \addplot [gray,mark=*,mark size=1pt] table [x=idx, y=rand] {data.dat};
        \addplot [cyan,mark=.,mark size=1pt] table [x=idx, y=bald] {data.dat};
        \addplot [violet,mark=x,mark size=1pt] table [x=idx, y=balddpp] {data.dat};
        \addplot [cyan,mark=.,mark size=1pt, dotted, thick, mark options=solid] table [x=idx, y=balddup] {data.dat};
        \addplot [violet,mark=x,mark size=1pt, dotted, mark options=solid] table [x=idx, y=balddppdup] {data.dat};

        \addplot [name path=upper,draw=none] table[x=idx,y expr=\thisrow{rand}+\thisrow{randerr}] {data.dat};
        \addplot [name path=lower,draw=none] table[x=idx,y expr=\thisrow{rand}-\thisrow{randerr}] {data.dat};
        \addplot [fill=gray!20] fill between[of=upper and lower];

        \addplot [name path=upper,draw=none] table[x=idx,y expr=\thisrow{bald}+\thisrow{balderr}] {data.dat};
        \addplot [name path=lower,draw=none] table[x=idx,y expr=\thisrow{bald}-\thisrow{balderr}] {data.dat};
        \addplot [fill=cyan!10] fill between[of=upper and lower];

        \addplot [name path=upper,draw=none] table[x=idx,y expr=\thisrow{balddpp}+\thisrow{balddpperr}] {data.dat};
        \addplot [name path=lower,draw=none] table[x=idx,y expr=\thisrow{balddpp}-\thisrow{balddpperr}] {data.dat};
        \addplot [fill=violet!10] fill between[of=upper and lower];

        \addplot [name path=upper,draw=none] table[x=idx,y expr=\thisrow{balddup}+\thisrow{baldduperr}] {data.dat};
        \addplot [name path=lower,draw=none] table[x=idx,y expr=\thisrow{balddup}-\thisrow{baldduperr}] {data.dat};
        \addplot [fill=cyan!10] fill between[of=upper and lower];

        \legend{Random,BALD,BALD w/ DPP,BALD (dup),BALD w/ DPP (dup)}

        \end{axis}
    \end{tikzpicture}
}
\subfloat{
    \begin{tikzpicture}[]
        \begin{axis}[
            width=0.37*\textwidth,
            grid=none,
            xmin=1,
            xmax=32,
            ymin=82,
            ymax=92,
            legend pos=south east,
            legend style={draw=none, font=\tiny, row sep=-1pt},
            yticklabel style = {font=\tiny},
            xticklabel style = {font=\tiny},
        ]

        \addplot [gray,mark=*,mark size=1pt] table [x=idx, y=rand] {data.dat};
        \addplot [cyan,mark=.,mark size=1pt] table [x=idx, y=id] {data.dat};
        \addplot [violet,mark=x,mark size=1pt] table [x=idx, y=iddpp] {data.dat};
        \addplot [cyan,mark=.,mark size=1pt, dotted, thick, mark options=solid] table [x=idx, y=iddup] {data.dat};
        \addplot [violet,mark=x,mark size=1pt, dotted, mark options=solid] table [x=idx, y=iddppdup] {data.dat};

        \addplot [name path=upper,draw=none] table[x=idx,y expr=\thisrow{rand}+\thisrow{randerr}] {data.dat};
        \addplot [name path=lower,draw=none] table[x=idx,y expr=\thisrow{rand}-\thisrow{randerr}] {data.dat};
        \addplot [fill=gray!20] fill between[of=upper and lower];

        \addplot [name path=upper,draw=none] table[x=idx,y expr=\thisrow{id}+\thisrow{iderr}] {data.dat};
        \addplot [name path=lower,draw=none] table[x=idx,y expr=\thisrow{id}-\thisrow{iderr}] {data.dat};
        \addplot [fill=cyan!10] fill between[of=upper and lower];

        \addplot [name path=upper,draw=none] table[x=idx,y expr=\thisrow{iddpp}+\thisrow{iddpperr}] {data.dat};
        \addplot [name path=lower,draw=none] table[x=idx,y expr=\thisrow{iddpp}-\thisrow{iddpperr}] {data.dat};
        \addplot [fill=violet!10] fill between[of=upper and lower];

        \addplot [name path=upper,draw=none] table[x=idx,y expr=\thisrow{iddup}+\thisrow{idduperr}] {data.dat};
        \addplot [name path=lower,draw=none] table[x=idx,y expr=\thisrow{iddup}-\thisrow{idduperr}] {data.dat};
        \addplot [fill=cyan!10] fill between[of=upper and lower];

        \legend{Random,ID,ID w/ DPP,ID (dup),ID w/ DPP (dup)}

        \end{axis}
    \end{tikzpicture}
}
\caption{
    Learning curves of different sampling strategies based on AMP (left), BALD (middle) and ID (right),
    comparing diversity-aware (w/ DPP) and diversity-agnostic variants
    using the original and duplicated corpus (dup).
    The $x$-axis shows the number of rounds for annotation.
    Random (dup) curves overlap with those of Random and
    are omitted for readability.
}
\label{fig:main-fig}
\end{figure*}
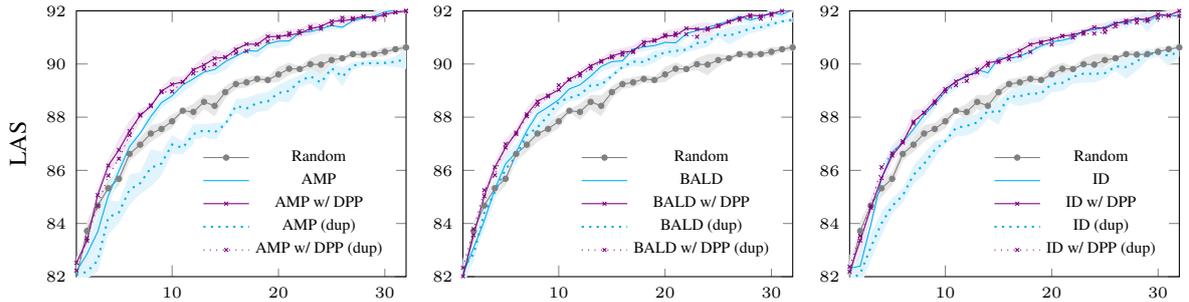

In our qualitative analysis (Appendix \ref{app:examples}),
we find that diversity-agnostic selection strategies tend to
select near-duplicate sentences.
To examine this phenomenon in isolation,
we repeat the training corpus twice
and observe the effect of diversity-aware strategies.
The corpus duplication technique has been previously used to probe
semantic models \citep{schofield+17}.
\reffig{fig:main-fig} shows learning curves for strategies
under the original and corpus duplication settings.
As expected, diversity-aware strategies consistently
outperform their diversity-agnostic counterparts
across both settings,
while some diversity-agnostic strategies (e.g., AMP)
even underperform uniform random selection
in the duplicated setting.

\paragraph{Interpreting the Effectiveness of Diversity-Agnostic Models}

\begin{figure}
\centering
\includegraphics[width=0.46\textwidth]{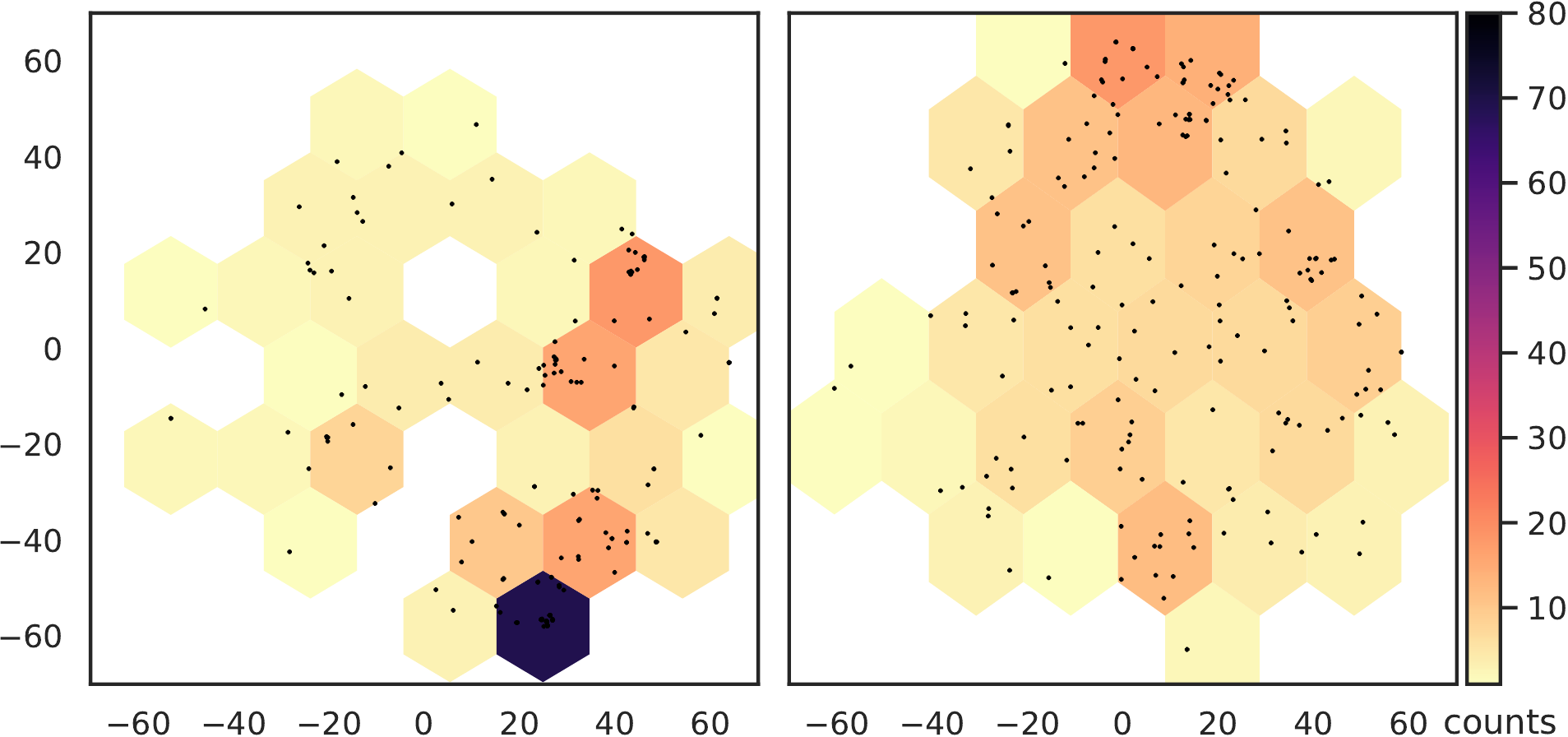}
\caption{
    t-SNE visualization of the distributions of the $200$ highest-quality
    unlabeled sentences over the diversity feature space
    after the 1\textsuperscript{st} (left) and the 10\textsuperscript{th} (right)
    annotation rounds
    using AMP without DPPs.
    Darker region indicates
    more data points residing in that diversity feature neighborhood.
    The left figure contains a dense region,
    while the data in the right figure are spread out in the feature space.
}
\label{fig:vis}
\end{figure}

\reffig{fig:vis} visualizes the density distributions of
the top $200$ data instances by AMP over the diversity feature space
reduced to two dimensions through t-SNE \citep{vandermaaten-hinton08}.
During the initial stage of active learning, data with the highest quality measures are
concentrated within a small neighborhood.
A diversity-agnostic strategy will sample similar examples for annotation.
After a few rounds of annotation and model training, the distribution of high quality examples spreads out,
and an AMP selection strategy is likely to sample a diverse set of examples without explicitly modeling diversity.
Our analysis corroborates previous findings \citep{thompson+99} that
small annotation batches are effective early in uncertainty sampling,
avoiding selecting many near-duplicate examples when intra-batch diversity is low,
but a larger batch size is more efficient later in training once intra-batch diversity increases.
\section{Related Work}
\label{sec:related_work}

Modeling diversity in batch-mode AL \citep{brinker2003incorporating}
has recently attracted attention in the machine learning community.
\citet{kirsch+19} introduce
a Bayesian batch-mode selection strategy
by estimating the mutual information
between a set of samples and the model parameters.
\citet{ash2020deep} present a
diversity-inducing sampling method
using gradient embeddings.
Most related to our work,
\citet{biyik2019batch} first apply DPPs to batch-mode AL.
Building on their approach, we flesh out a DPP treatment for AL for
a structured prediction task, dependency parsing.
Previously, \citet{shen2018deep} consider named entity recognition
but they report negative results for a diversity-inducing variant
of their sampling method.

Due to the high annotation cost,
AL is a popular technique for parsing
and parse selection \citep{osborne-baldridge04}.
Recent advances
focus on reducing full-sentence annotations
to a subset of tokens within a sentence
\citep{sassano-kurohashi10,mirroshandel-nasr11,majidi-crane13a,flannery-mori15,li2016active}.
We show that AL for parsing
can further benefit from
diversity-aware sampling strategies.

DPPs have previously been successfully applied to
the tasks of extractive text summarization \citep{cho+19a,cho+19} and
modeling phoneme inventories \citep{cotterell-eisner17}.
In this work, we show that DPPs
also provide a useful framework for understanding and modeling
quality and diversity in active learning for NLP tasks.

\section{Conclusion}
\label{sec:conclusion}

We show that compared with their diversity-agnostic counterparts,
diversity-aware sampling strategies
not only lead to higher data efficiency,
but are also more robust under corpus duplication settings.
Our work invites future research into
methods, utility and success conditions for
modeling diversity
in active learning for
NLP tasks.

\section*{Acknowledgements}

We thank the anonymous reviewers for their insightful reviews,
and Prabhanjan Kambadur,
Chen-Tse Tsai,
and Minjie Xu
for discussion and comments.
Tianze Shi acknowledges support from Bloomberg's Data Science Ph.D. Fellowship.

\bibliography{main}
\bibliographystyle{acl_natbib}

\clearpage

\begin{appendices}
\setcounter{table}{0}
\setcounter{figure}{0}
\setcounter{algocf}{0}
\renewcommand{\thetable}{\Alph{section}\arabic{table}}
\renewcommand{\thefigure}{\Alph{section}\arabic{figure}}
\renewcommand{\thealgocf}{\Alph{section}\arabic{algocf}}
\section{Dependency Parser}
\label{app:parser}

We adopt the deep biaffine dependency parser proposed by \citet{dozat-manning17}.
The parser is conceptually simple and yet competitive with
state-of-the-art dependency parsers.
The parser has three components:
feature extraction, unlabeled parsing and relation labeler.

\paragraph{Feature Extraction}
For a length-$n$ sentence $x=x_0, x_1, x_2, \ldots, x_n$,
where $x_0$ is the dummy root symbol,
we extract contextualized features at each word position.
In our experiments, we use a pretrained multilingual XLM-RoBERTa base model \citep{conneau+20},
and fine-tune the feature extractor along with the rest of our parser:
$$
[\vect{x}_0, \vect{x}_1,\ldots, \vect{x}_n] = \text{XLM-R}(x_0, x_1, \ldots, x_n).
$$
Each word input to the XLM-RoBERTa model is
processed with the SentencePiece tokenizer \citep{kudo-richardson18},
and we follow \citet{kitaev+19} and retain
the vectors corresponding to the last sub-word units as their representations.
For~$\vect{x}_0$, we use the vector of the \texttt{[CLS]} token,
which is appended by XLM-RoBERTa to the beginning of each sentence.

\paragraph{Unlabeled Parser}
The parser uses a deep biaffine attention mechanism to derive
locally-normalized attachment probabilities for all potential head-dependent pairs:
\begin{align*}
\vect{h}_i^{\text{arc-head}}&=\text{MLP}^{\text{arc-head}}(\vect{x}_i)\\
\vect{h}_j^{\text{arc-dep}}&=\text{MLP}^{\text{arc-dep}}(\vect{x}_j)\\
s_{i,j} &= [\vect{h}_i^\text{arc-head};1]^\top U^\text{arc} [\vect{h}_j^\text{arc-dep};1] \\
P_\text{att}(\text{head(j)}=i~|~x) &= \text{softmax}_i(s_{:,j}),
\end{align*}
where $\text{MLP}^\text{arc-head}$ and $\text{MLP}^\text{arc-dep}$
are two multi-layer perceptrons (MLPs) projecting $\vect{x}$ vectors
into $d^\text{arc}$-dimensional $\vect{h}$ vectors,
$[;1]$ appends an element of $1$ at the end of the vectors,
and $U^\text{arc}\in \mathbb{R}^{(d^\text{arc}+1)\times (d^\text{arc}+1)}$ is a bilinear scoring matrix.
This component is trained with cross-entropy loss of the gold-standard attachments.
During inference, we use the Chu-Liu-Edmonds algorithm \citep{chu-liu65,edmonds67} to find the spanning tree
with the highest product of locally-normalized attachment probabilities.

\paragraph{Relation Labeler}

The relation labeling component employs a similar
deep biaffine scoring function as
the unlabeled parsing component:
\begin{align*}
\vect{h}_i^{\text{rel-head}}&=\text{MLP}^{\text{rel-head}}(\vect{x}_i)\\
\vect{h}_j^{\text{rel-dep}}&=\text{MLP}^{\text{rel-dep}}(\vect{x}_j)\\
t_{i,j,r} &= [\vect{h}_i^\text{rel-head};1]^\top U^\text{rel}_r [\vect{h}_j^\text{rel-dep};1] \\
P(\text{rel}(i, j)=r) &= \text{softmax}_r(t_{i,j,:}),
\end{align*}
where each $U^\text{rel}_r\in \mathbb{R}^{(d^\text{rel}+1)\times(d^\text{rel}+1)}$,
and there are as many such matrices as the size of
the dependency relation label set $|R|$.
The relation labeler is trained using cross entropy loss
on the gold-standard head-dependent pairs. During inference,
the labeling decision for each arc is
made independently given
the predicted unlabeled parse tree.

\section{Results with UAS Evaluation}
\label{app:uas}

\begin{table}[]
    \small
    \begin{tabular}{@{\hspace{3pt}}l@{\hspace{2pt}}|c@{\hspace{3pt}}c|c@{\hspace{3pt}}c@{\hspace{3pt}}}
    \toprule
     Round \# & \multicolumn{2}{c|}{$5$} & \multicolumn{2}{c}{$10$} \\
    \midrule
     Strategy & w/o DPP   & w/ DPP   & w/o DPP   & w/ DPP   \\
    \midrule
     Random & \numstd{89.01}{.28}  & \numstdbf{89.67}{.30}
            & \numstd{90.78}{.27}  & \numstdbf{91.22}{.22} \\
     AMP    & \numstd{89.67}{.29}  & \numstdbf{90.24}{.39}
            & \numstd{92.03}{.10}  & \numstdbf{92.17}{.22} \\
     BALD   & \numstd{89.82}{.36}  & \numstdbf{90.29}{.20}
            & \numstd{91.87}{.36}  & \numstdbf{92.00}{.08} \\
     ID     & \numstdbf{90.24}{.20}  & \numstd{90.03}{.18}
            & \numstdbf{92.16}{.17}  & \numstd{92.06}{.16} \\
    \bottomrule
    \end{tabular}
    \caption{UAS after $5$ and $10$ rounds of annotation
    (roughly $5{,}000$ and $7{,}000$ training tokens respectively),
    comparing strategies with and without modeling diversity through DPP.}
    \label{tab:uas-results}
\end{table}
We also evaluate different learning strategies
based on unlabeled attachment scores (UAS),
and the results are shown in \reftab{tab:uas-results}.
In line with LAS-based experiments, we find that modeling diversity is more helpful during initial stages of learning.
For ID, we observe this effect even earlier than the fifth round of annotation:
\numstd{86.57}{.44} vs. \numstd{87.40}{.51} after the first annotation round.

\section{Sentence Selection Examples}
\label{app:examples}

\begin{table*}[]
    \tiny
    \begin{tabular}{cp{14.8cm}}
    \toprule
    & {\bf \small Sentences selected by AMP (highest-quality ones first):}\\\vspace{1pt}
    & Downgraded by Moody 's were Houston Lighting 's first - mortgage bonds and secured pollution - control bonds to single - A - 3 from single - A - 2 ; unsecured pollution - control bonds to Baa - 1 from single - A - 3 ; preferred stock to single - A - 3 from single - A - 2 ; a shelf registration for preferred stock to a preliminary rating of single - A - 3 from a preliminary rating of single - A - 2 ; two shelf registrations for collateralized debt securities to a preliminary rating of single - A - 3 from a preliminary rating of single - A - 2 , and the unit 's rating for commercial paper to Prime - 2 from Prime - 1 .\\
    & For a while in the 1970s it seemed Mr. Moon was on a spending spree , with such purchases as the former New Yorker Hotel and its adjacent Manhattan Center ; a fishing / processing conglomerate with branches in Alaska , Massachusetts , Virginia and Louisiana ; a former Christian Brothers monastery and the Seagram family mansion ( both picturesquely situated on the Hudson River ) ; shares in banks from Washington to Uruguay ; a motion picture production company , and newspapers , such as the Washington Times , the New York City Tribune ( originally the News World ) , and the successful Spanish - language Noticias del Mundo .\\
    $\rightarrow$ & \textcolor{Mahogany}{LONDON LATE EURODOLLARS : 8 11/16 \% to 8 9/16 \% one month ; 8 5/8 \% to 8 1/2 \% two months ; 8 5/8 \% to 8 1/2 \% three months ; 8 9/16 \% to 8 7/16 \% four months ; 8 1/2 \% to 8 3/8 \% five months ; 8 1/2 \% to 8 3/8 \% six months .}\\
    $\rightarrow$ & \textcolor{Mahogany}{LONDON LATE EURODOLLARS : 8 3/4 \% to 8 5/8 \% one month ; 8 3/4 \% to 8 5/8 \% two months ; 8 11/16 \% to 8 9/16 \% three months ; 8 9/16 \% to 8 7/16 \% four months ; 8 1/2 \% to 8 3/8 \% five months ; 8 7/16 \% to 8 5/16 \% six months .}\\
    $\rhd$ & \textcolor{Blue}{COMMERCIAL PAPER placed directly by General Motors Acceptance Corp. : 8.40 \% 30 to 44 days ; 8.325 \% 45 to 59 days ; 8.10 \% 60 to 89 days ; 8 \% 90 to 119 days ; 7.85 \% 120 to 149 days ; 7.70 \% 150 to 179 days ; 7.375 \% 180 to 270 days .}\\
    & 4 . When a RICO TRO is being sought , the prosecutor is required , at the earliest appropriate time , to state publicly that the government 's request for a TRO , and eventual forfeiture , is made in full recognition of the rights of third parties -- that is , in requesting the TRO , the government will not seek to disrupt the normal , legitimate business activities of the defendant ; will not seek through use of the relation - back doctrine to take from third parties assets legitimately transferred to them ; will not seek to vitiate legitimate business transactions occurring between the defendant and third parties ; and will , in all other respects , assist the court in ensuring that the rights of third parties are protected , through proceeding under { RICO } and otherwise .\\
    $\rhd$ & \textcolor{Blue}{COMMERCIAL PAPER placed directly by General Motors Acceptance Corp. : 8.50 \% 30 to 44 days ; 8.25 \% 45 to 62 days ; 8.375 \% 63 to 89 days ; 8 \% 90 to 119 days ; 7.90 \% 120 to 149 days ; 7.80 \% 150 to 179 days ; 7.55 \% 180 to 270 days .}\\
    $\rhd$ & \textcolor{Blue}{COMMERCIAL PAPER placed directly by General Motors Acceptance Corp. : 8.50 \% 30 to 44 days ; 8.25 \% 45 to 65 days ; 8.375 \% 66 to 89 days ; 8 \% 90 to 119 days ; 7.875 \% 120 to 149 days ; 7.75 \% 150 to 179 days ; 7.50 \% 180 to 270 days .}\\
    $\rightarrow$ & \textcolor{Mahogany}{LONDON LATE EURODOLLARS : 8 11/16 \% to 8 9/16 \% one month ; 8 5/8 \% to 8 1/2 \% two months ; 8 5/8 \% to 8 1/2 \% three months ; 8 9/16 \% to 8 7/16 \% four months ; 8 1/2 \% to 8 3/8 \% five months ; 8 7/16 \% to 8 5/16 \% six months .}\\
    $\rightarrow$ & \textcolor{Mahogany}{LONDON LATE EURODOLLARS : 8 11/16 \% to 8 9/16 \% one month ; 8 9/16 \% to 8 7/16 \% two months ; 8 5/8 \% to 8 1/2 \% three months ; 8 1/2 \% to 8 3/8 \% four months ; 8 7/16 \% to 8 5/16 \% five months ; 8 7/16 \% to 8 5/16 \% six months .}\\
    & The new edition lists the top 10 metropolitan areas as Anaheim - Santa Ana , Calif. ; Boston ; Louisville , Ky. ; Nassau - Suffolk , N.Y. ; New York ; Pittsburgh ; San Diego ; San Francisco ; Seattle ; and Washington .\\
    $\rhd$ & \textcolor{Blue}{COMMERCIAL PAPER placed directly by General Motors Acceptance Corp. : 8.45 \% 30 to 44 days ; 8.20 \% 45 to 67 days ; 8.325 \% 68 to 89 days ; 8 \% 90 to 119 days ; 7.875 \% 120 to 149 days ; 7.75 \% 150 to 179 days ; 7.50 \% 180 to 270 days .}\\
    $\rhd$ & \textcolor{Blue}{COMMERCIAL PAPER placed directly by General Motors Acceptance Corp. : 8.50 \% 2 to 44 days ; 8.25 \% 45 to 69 days ; 8.40 \% 70 to 89 days ; 8.20 \% 90 to 119 days ; 8.05 \% 120 to 149 days ; 7.90 \% 150 to 179 days ; 7.50 \% 180 to 270 days .}\\
    & Five officials of this investment banking firm were elected directors : E. Garrett Bewkes III , a 38 - year - old managing director in the mergers and acquisitions department ; Michael R. Dabney , 44 , a managing director who directs the principal activities group which provides funding for leveraged acquisitions ; Richard Harriton , 53 , a general partner who heads the correspondent clearing services ; Michael Minikes , 46 , a general partner who is treasurer ; and William J. Montgoris , 42 , a general partner who is also senior vice president of finance and chief financial officer .\\
    $\rightarrow$ & \textcolor{Mahogany}{LONDON LATE EURODOLLARS : 8 11/16 \% to 8 9/16 \% one month ; 8 5/8 \% to 8 1/2 \% two months ; 8 11/16 \% to 8 9/16 \% three months ; 8 9/16 \% to 8 7/16 \% four months ; 8 1/2 \% to 8 3/8 \% five months ; 8 7/16 \% to 8 5/16 \% six months .}\\
    $\rhd$ & \textcolor{Blue}{COMMERCIAL PAPER placed directly by General Motors Acceptance Corp. : 8.55 \% 30 to 44 days ; 8.25 \% 45 to 59 days ; 8.40 \% 60 to 89 days ; 8 \% 90 to 119 days ; 7.90 \% 120 to 149 days ; 7.80 \% 150 to 179 days ; 7.55 \% 180 to 270 days .}\\
    & They transferred some \$ 28 million from the Community Development Block Grant program designated largely for low - and moderate - income projects and funneled it into such items as : -- \$ 1.2 million for a performing - arts center in Newark , -- \$ 1.3 million for `` job retention '' in Hawaiian sugar mills . -- \$ 400,000 for a collapsing utility tunnel in Salisbury , -- \$ 500,000 for `` equipment and landscaping to deter crime and aid police surveillance '' at a Michigan park . -- \$ 450,000 for `` integrated urban data based in seven cities . '' No other details . -- \$ 390,000 for a library and recreation center at Mackinac Island , Mich .\\
    $\rightarrow$ & \textcolor{Mahogany}{LONDON LATE EURODOLLARS : 8 3/4 \% to 8 5/8 \% one month ; 8 13/16 \% to 8 11/16 \% two months ; 8 11/16 \% to 8 9/16 \% three months ; 8 9/16 \% to 8 7/16 \% four months ; 8 1/2 \% to 8 3/8 \% five months ; 8 7/16 \% to 8 5/16 \% six months .}\\
    $\rhd$ & \textcolor{Blue}{COMMERCIAL PAPER placed directly by General Motors Acceptance Corp. : 8.45 \% 30 to 44 days ; 8.25 \% 45 to 68 days ; 8.30 \% 69 to 89 days ; 8.125 \% 90 to 119 days ; 8 \% 120 to 149 days ; 7.875 \% 150 to 179 days ; 7.50 \% 180 to 270 days .}\\
    \midrule
    & {\bf \small Sentences selected by AMP with diversity-inducing DPP:}\\\vspace{0pt}
    & Downgraded by Moody 's were Houston Lighting 's first - mortgage bonds and secured pollution - control bonds to single - A - 3 from single - A - 2 ; unsecured pollution - control bonds to Baa - 1 from single - A - 3 ; preferred stock to single - A - 3 from single - A - 2 ; a shelf registration for preferred stock to a preliminary rating of single - A - 3 from a preliminary rating of single - A - 2 ; two shelf registrations for collateralized debt securities to a preliminary rating of single - A - 3 from a preliminary rating of single - A - 2 , and the unit 's rating for commercial paper to Prime - 2 from Prime - 1 .\\
    & 4 . When a RICO TRO is being sought , the prosecutor is required , at the earliest appropriate time , to state publicly that the government 's request for a TRO , and eventual forfeiture , is made in full recognition of the rights of third parties -- that is , in requesting the TRO , the government will not seek to disrupt the normal , legitimate business activities of the defendant ; will not seek through use of the relation - back doctrine to take from third parties assets legitimately transferred to them ; will not seek to vitiate legitimate business transactions occurring between the defendant and third parties ; and will , in all other respects , assist the court in ensuring that the rights of third parties are protected , through proceeding under { RICO } and otherwise .\\
    $\rhd$ & \textcolor{Blue}{COMMERCIAL PAPER placed directly by General Motors Acceptance Corp. : 8.40 \% 30 to 44 days ; 8.325 \% 45 to 59 days ; 8.10 \% 60 to 89 days ; 8 \% 90 to 119 days ; 7.85 \% 120 to 149 days ; 7.70 \% 150 to 179 days ; 7.375 \% 180 to 270 days .}\\
    & Moreover , the process is n't without its headaches .\\
    & For a while in the 1970s it seemed Mr. Moon was on a spending spree , with such purchases as the former New Yorker Hotel and its adjacent Manhattan Center ; a fishing / processing conglomerate with branches in Alaska , Massachusetts , Virginia and Louisiana ; a former Christian Brothers monastery and the Seagram family mansion ( both picturesquely situated on the Hudson River ) ; shares in banks from Washington to Uruguay ; a motion picture production company , and newspapers , such as the Washington Times , the New York City Tribune ( originally the News World ) , and the successful Spanish - language Noticias del Mundo .\\
    & Within the paper sector , Mead climbed 2 3/8 to 38 3/4 on 1.3 million shares , Union Camp rose 2 3/4 to 37 3/4 , Federal Paper Board added 1 3/4 to 23 7/8 , Bowater gained 1 1/2 to 27 1/2 , Stone Container rose 1 to 26 1/8 and Temple - Inland jumped 3 3/4 to 62 1/4 .\\
    & We finally rendezvoused with our balloon , which had come to rest on a dirt road amid a clutch of Epinalers who watched us disassemble our craft -- another half - an - hour of non-flight activity -- that included the precision routine of yanking the balloon to the ground , punching all the air out of it , rolling it up and cramming it and the basket into the trailer .\\
    & These are the 26 states , including the commonwealth of Puerto Rico , that have settled with Drexel : Alaska , Arkansas , Delaware , Georgia , Hawaii , Idaho , Indiana , Iowa , Kansas , Kentucky , Maine , Maryland , Minnesota , Mississippi , New Hampshire , New Mexico , North Dakota , Oklahoma , Oregon , South Carolina , South Dakota , Utah , Vermont , Washington , Wyoming and Puerto Rico .\\
    & It is the stuff of dreams , but also of traumas .\\
    & An inquiry into his handling of Lincoln S\&L inevitably will drag in Sen. Cranston and the four others , Sens. Dennis DeConcini ( D. , Ariz. ) , John McCain ( R. , Ariz. ) , John Glenn ( D. , Ohio ) and Donald Riegle ( D. , Mich . ) .\\
    & Five officials of this investment banking firm were elected directors : E. Garrett Bewkes III , a 38 - year - old managing director in the mergers and acquisitions department ; Michael R. Dabney , 44 , a managing director who directs the principal activities group which provides funding for leveraged acquisitions ; Richard Harriton , 53 , a general partner who heads the correspondent clearing services ; Michael Minikes , 46 , a general partner who is treasurer ; and William J. Montgoris , 42 , a general partner who is also senior vice president of finance and chief financial officer .\\
    & But as they hurl fireballs that smolder rather than burn , and relive old duels in the sun , it 's clear that most are there to make their fans cheer again or recapture the camaraderie of seasons past or prove to themselves and their colleagues that they still have it -- or something close to it .\\
    & They are : `` A Payroll to Meet : A Story of Greed , Corruption and Football at SMU '' ( Macmillan , 221 pages , \$ 18.95 ) by David Whitford ; `` Big Red Confidential : Inside Nebraska Football '' ( Contemporary , 231 pages , \$ 17.95 ) by Armen Keteyian ; and `` Never Too Young to Die : The Death of Len Bias '' ( Pantheon , 252 pages , \$ 18.95 ) by Lewis Cole .\\
    & He says he told NewsEdge to look for stories containing such words as takeover , acquisition , acquire , LBO , tender , merger , junk and halted .\\
    & It is no coincidence that from 1844 to 1914 , when the Bank of England was an independent private bank , the pound was never devalued and payment of gold for pound notes was never suspended , but with the subsequent nationalization of the Bank of England , the pound was devalued with increasing frequency and its use as an international medium of exchange declined .\\
    & The \$ 4 billion in bonds break down as follows : \$ 1 billion in five - year bonds with a coupon rate of 8.25 \% and a yield to maturity of 8.33 \% ; \$ 1 billion in 10 - year bonds with a coupon rate of 8.375 \% and a yield to maturity of 8.42 \% ; \$ 2 billion in 30 - year bonds with five - year call protection , a coupon rate of 8.75 \% and a yield to maturity of 9.06 \% .\\
    & Hecla Mining rose 5/8 to 14 ; Battle Mountain Gold climbed 3/4 to 16 3/4 ; Homestake Mining rose 1 1/8 to 16 7/8 ; Lac Minerals added 5/8 to 11 ; Placer Dome went up 7/8 to 16 3/4 , and ASA Ltd. jumped 3 5/8 to 49 5/8 .\\
    \bottomrule
    \end{tabular}
    \caption{
    Sentences picked by a diversity-agnostic (top)
    and a diversity-aware (bottom) selection strategy
    from the same unlabeled pool
    after the intial round of model training on the seed sentences.
    The diversity-agnostic strategy selects many near-duplicate sentences
    (the two near-duplicate clusters are marked by red $\rightarrow$ and blue $\rhd$),
    effectively wasting the annotation budget,
    where DPPs largely alleviate this issue by enforcing diversity.
    }
    \label{tab:sample-batches}
\end{table*}

In \reftab{tab:sample-batches} we compare batches sampled by a diversity-aware selection strategy with a diversity-agnostic one.
We observe that by modeling diversity in the sample selection process,
DPPs avoid selecting duplicate or near-duplicate sentences
and thus the annotation budget can be maximally utilized.

\section{BALD under High Duplication Setting}
\label{app:bald}

\begin{figure}
    \centering
    \begin{tikzpicture}[]
        \begin{axis}[
            width=0.38*\textwidth,
            grid=none,
            ylabel=LAS,
            xlabel=Annotation round,
            xmin=1,
            xmax=32,
            ymin=82,
            ymax=92,
            legend pos=south east,
            legend style={draw=none, font=\tiny, row sep=-1pt},
            yticklabel style = {font=\tiny},
            xticklabel style = {font=\tiny},
            ylabel style = {font=\small},
            xlabel style = {font=\small},
        ]

        \addplot [gray,mark=*,mark size=1pt] table [x=idx, y=random] {data-bald.dat};
        \addplot [cyan,mark=.,mark size=1pt] table [x=idx, y=bald] {data-bald.dat};
        \addplot [violet,mark=x,mark size=1pt] table [x=idx, y=balddpp] {data-bald.dat};

        \addplot [name path=upper,draw=none] table[x=idx,y expr=\thisrow{random}+\thisrow{randomerr}] {data-bald.dat};
        \addplot [name path=lower,draw=none] table[x=idx,y expr=\thisrow{random}-\thisrow{randomerr}] {data-bald.dat};
        \addplot [fill=gray!20] fill between[of=upper and lower];

        \addplot [name path=upper,draw=none] table[x=idx,y expr=\thisrow{bald}+\thisrow{balderr}] {data-bald.dat};
        \addplot [name path=lower,draw=none] table[x=idx,y expr=\thisrow{bald}-\thisrow{balderr}] {data-bald.dat};
        \addplot [fill=cyan!10] fill between[of=upper and lower];

        \addplot [name path=upper,draw=none] table[x=idx,y expr=\thisrow{balddpp}+\thisrow{balddpperr}] {data-bald.dat};
        \addplot [name path=lower,draw=none] table[x=idx,y expr=\thisrow{balddpp}-\thisrow{balddpperr}] {data-bald.dat};
        \addplot [fill=violet!10] fill between[of=upper and lower];

        \legend{Random, BALD (dup), BALD w/ DPP (dup)}

        \end{axis}
    \end{tikzpicture}
\caption{
    Learning curves for BALD-based selection strategies
    under a five-fold corpus duplication setting.
}
\label{fig:bald}
\end{figure}
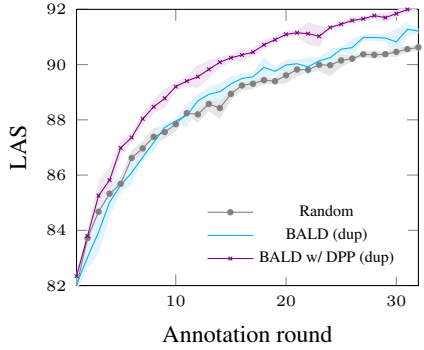

\reffig{fig:bald} shows the learning curves for BALD-based selection strategies
under a high corpus duplication setting
where the corpus is repeated five times.
In this extreme setting, the diversity-agnostic strategy significantly underperforms the diversity-aware one.
We posit that the relative success of BALD compared to AMP
in the twice-duplicated setting is due to the fact that
BALD randomly draws dropout masks to estimate model uncertainty,
so that identical examples could still have
different quality measures.

\section{Implementation Details and Hyperparameters}
\label{app:hyperparameters}

We do not tune our hyperparameters
since in practice, active learning systems only
have a single shot at success, without tuning.
Instead, we follow recommendations from
relevant prior work in setting our learning details and hyperparameters.

\begin{algorithm}[t]
\small
\SetAlgoLined
\KwIn{candidate item set $X$ (sentences or tokens),
DPP represented by matrix~$L$, size budget $b$}
$U\gets X$\;
$Y\gets \emptyset$\;
\While{$U\neq \emptyset$}{
$i \gets \arg\max_{i'\in U}\text{det}(L_{Y\cup \{i'\}})$\;
\eIf{\upshape $\sum_{y\in Y}\text{size}(y)< b$}
{$Y\gets Y \cup \{i\}$\;}
{break\;}
}

\KwOut{selected items $Y$}

\caption{Greedy MAP inference for DPP with a size budget, adapted from \citet{kulesza2012thesis}.}
\label{algo:dpp-map}

\end{algorithm}

\paragraph{Active Learning}

Following \citet{li2016active}, our active learning set-up
proceeds in two stages for each annotation round.
In the first stage, we select sentences filling in a budget of $2500$ tokens;
in the second stage, we pick $500$ tokens out of the subset of sentences.
For a diversity-agnostic strategy,
we choose the top-$k$ highest-quality candidates within the token budget,
while our diversity-aware selection strategy uses a separate DPP for each stage.
The active learning process is bootstraped with a seed set of $128$ labeled sentences.
For the BALD quality measure, we set $K=5$.

\paragraph{Greedy MAP Inference for DPPs}

\refalgo{algo:dpp-map} illustrates the procedure for
selecting items from DPPs under a budget constraint.
This greedy MAP inference algorithm
is adapted from \citet{kulesza2012thesis}.
During sentence selection, the size of a sentence is its number of tokens,
and each token has a size of $1$ in the token selection stage.

\paragraph{Dependency Parser}
We set the hyperparameters according to \citet{dozat-manning17}.
All the MLPs in the deep biaffine attention architecture have single hidden layers
with ReLU activation functions and a dropout probability of $0.33$,
and we set $d^\text{arc}$ and $d^\text{rel}$ to be $500$ and $100$ respectively.

\paragraph{Training and Optimization}

Each training batch contains $16$ sentences
and gradient norms are clipped to $5.0$.
We use the Adam optimizer \citep{kingma-ba15} with a learning rate of $10^{-5}$
with $640$ warmup steps with a linearly-increasing learning rate starting from $0$.

\paragraph{Implementation}
Our implementation is in PyTorch \cite{paszke+19},
and we use the \texttt{transformers} package\footnote{\url{https://github.com/huggingface/transformers}}
to interface with the pretrained XLM-RoBERTa model.
\end{appendices}

\end{document}